\documentclass[conference]{IEEEtran}
\IEEEoverridecommandlockouts
\usepackage[utf8]{inputenc}
\usepackage[english]{babel}
\usepackage{cite}
\usepackage{amsmath,amssymb,amsfonts}
\usepackage{booktabs, cellspace, hhline}
\usepackage{algorithmic}
\usepackage{graphicx}
\usepackage{textcomp}
\usepackage{xcolor}
\usepackage{listings}
\usepackage{hyperref}
\usepackage{wrapfig}
\usepackage[ampersand]{easylist}
\usepackage{color}
\usepackage{scrextend}
\usepackage{amsmath}
\usepackage{multirow}
\usepackage{array}
\usepackage{subcaption}
\usepackage[export, demo]{adjustbox}
\usepackage{cellspace, tabularx}
\usepackage{mwe}
\usepackage{graphbox}
\usepackage{ragged2e}
\usepackage[labelfont=bf]{caption}
\graphicspath{ {./images/} }

\begin{document}

\title{\textbf{A Novel Explainable Artificial Intelligence Model in Image Classification problem}\\
}
\makeatletter
\newcommand{\linebreakand}{%
  \end{@IEEEauthorhalign}
  \hfill\mbox{}\par
  \hfill\begin{@IEEEauthorhalign}
}
\makeatother

\author{
\IEEEauthorblockN{Hung Quoc Cao}
\IEEEauthorblockA{\textit{FSO.QNH.QAI.AIC} \\
\textit{FPT Software}\\
Binh Dinh, Vietnam \\
HungCQ3@fsoft.com.vn}
\and
\IEEEauthorblockN{Hung Truong Thanh Nguyen\textsuperscript{\textbf{1, 2}}}
\IEEEauthorblockA{\textsuperscript{\textbf{1}}\textit{FSO.QNH.QAI.AIC} \\
\textit{FPT Software}\\
\textsuperscript{\textbf{2}}\textit{Department of Computer Science} \\
\textit{Frankfurt University of Applied Sciences}\\
Frankfurt am Main, Germany \\
HungNTT@fsoft.com.vn}
\\
\linebreakand
\IEEEauthorblockN{Khang Vo Thanh Nguyen}
\IEEEauthorblockA{\textit{FSO.QNH.QAI.AIC} \\
\textit{FPT Software}\\
Binh Dinh, Vietnam \\
KhangNVT1@fsoft.com.vn}
\and
\IEEEauthorblockN{Phong Xuan Nguyen}
\IEEEauthorblockA{\textit{Graduate School of Engineering Advanced Interdisciplinary Studies} \\
\textit{The University of Tokyo}\\
xphongvn@gmail.com}
}

\maketitle

\begin{abstract}
                    
In recent years, artificial intelligence is increasingly being applied widely in many different fields and has a profound and direct impact on human life. Following this is the need to understand the principles of the model making predictions. Since most of the current high-precision models are black boxes, neither the AI scientist nor the end-user deeply understands what's going on inside these models. Therefore, many algorithms are studied for the purpose of explaining AI models, especially those in the problem of image classification in the field of computer vision such as LIME, CAM, GradCAM. However, these algorithms still have limitations such as LIME's long execution time and CAM's confusing interpretation of concreteness and clarity. Therefore, in this paper, we propose a new method called Segmentation - Class Activation Mapping (SeCAM) that combines the advantages of these algorithms above, while at the same time overcoming their disadvantages.  We tested this algorithm with various models, including ResNet50, Inception-v3, VGG16 from ImageNet Large Scale Visual Recognition Challenge (ILSVRC) data set. Outstanding results when the algorithm has met all the requirements for a specific explanation in a remarkably concise time. 
\end{abstract}

\begin{IEEEkeywords}
Explainable Artificial Intelligence (XAI), machine learning, explanation, transparency, interpretability
\end{IEEEkeywords}

\section{Introduction}
In recent years, along with the rapid development of deep learning, more and more new models are being created with outstanding accuracy in the field of computer vision. However, those models have a complex network structure where users and scientists still cannot fully interpret and understand its black box \cite{mohseni2020multidisciplinary}. Arising from the need to explain to users and experts the reasons behind the model’s decisions or predictions, Explainable Artificial Intelligence (XAI) was born and is drawing more and more attention in AI. Various XAI methods have been introduced with different approaches, Adadi and Berrada \cite{XAI} have presented several ways to classify XAI algorithms,  it's based on scope, time of information extraction or model AI. With Scope-based classification, Global and Local are two variations according to the scope of interpretability. The Global XAI methods try to understand the entire model behavior while the Local XAI methods want to understand a single prediction. Global interpretability techniques lead users to trust a model. In reverse, local techniques lead users to believe in a prediction. Also, they try to identify each feature’s contributions in the input towards a particular output \cite{DBLP:journals/corr/abs-1808-00033}. Also for Based on Time of information extraction, we have two classes Post-hoc and Intrinsic. Post-hoc methods explain the model after being trained, these methods use model prediction and parameters to explain. Post-hoc methods can be applied to models without changing the model architecture. Therefore, Post-hoc methods can be applied to pretrained models. Intrinsic methods will modify the original model architecture. The model will be adjusted to have a new layer with interpretable constraints. And finally, model related methods, it’s another important way to classify XAI methods is whether they are agnostic or specific models. If an XAI method can be used for any type of model, it is classified as Agnostic. If an XAI method can be applied for only a single type or a number of classes of models, it is classified as Specific.

Although these methods can give satisfactory explanations, they still have many limitations and need to be improved. With the image classification problem, many XAI methods have been proposed, each method has a different approach and the output is therefore also different. For example: the output of LIME \cite{Lime} are the superpixels that affect the model's prediction most. SHAP \cite{Shap} will show the impact on the prediction, either positive or negative, of all superpixels. Those superpixels come from a previous perturbation step. With visualization algorithms like CAM \cite{Cam}, SISE \cite{sise}, Saliency Map \cite{saliencymap}... the output will show the user a heatmap on the original image.
From the knowledge we have gained through researching and comparing the three algorithms: LIME, SHAP, CAM. We recognize that LIME's explanation, the areas with the most influence, is the most intuitive and accurate for the image classification problem. These regions explained by LIME are approximately equivalent to how humans perceive the object. Nevertheless, the calculation time of LIME is too high. But, when we explain an image, the average time of LIME is 200 seconds greater than that of the CAM. However, the choice of the number of the most influenced regions is still dependent on people and specific image \cite{Lime}. The recent works have proposed a method to improve the computation speed of LIME, namely Modified Perturbed Sampling for LIME (MPS-LIME) [6]. In their experimental results with Google’s pre-trained Inception neural network on Image-net database, the runtime of MPS-LIME is nearly half as the runtime of LIME; but the calculation time is still incredibly long. In contrast, CAM does not suffer from these limitations, but the high impact areas of CAM are far broader than the human-defined bounding box. Moreover, CAM must modify the original model’s layers to work. In this work, we propose a new local post-hoc method of XAI in the image classification problem, called Segmentation - Class Activation Mapping (SeCAM). That method selects the regions that affect the model’s prediction as LIME, but with much faster time (approximate to CAM) . We believe that this concept of segmentation is also applicable to the class of CAM-based XAI methods \cite{sise}
Our main contribution are:
\begin{itemize}
    \item Propose SECAM as a new local post-hoc agnostic method of XAI in the image classification problem, which combines these advantages of the above two algorithms (LIME and CAM), and at the same time, overcomes their inherent weaknesses. Specifically, it can provide friendly images, close to human explanations like LIME while ensuring computation speed as fast as CAM, averaging 2 seconds for an explanation; Moreover, it also overcomes the weakness of having to edit the original model of the CAM method in some specific models.
    \item In addition to applying SeCAM to explain AI models in image classification problems, this approach’s main idea has much potential to improve other related XAI algorithms, especially in the computer vision field.
    \item We have experimented with datasets, models, ... and a number of qualitative as well as quantitative evaluation methods.
    \item We have a discussion about what it means to view an image as these superpixel to a user through a user study. We believe that, with the representation of an image in the form of superpixels, each superpixel will have some meaning, for example the head area, body area, ... of the object. Therefore, the user will be able to learn how the parts of the object affect the prediction of the model, or what do they mean?
    \item We also experiment with many segmentation algorithms to see how much impact it has on XAI algorithms such as LIME, SHAP, SeCAM - the algorithms use segmentation algorithms as part of the perturbation step.
    \item We had a survey with real users to see what the given explanations mean to them.
    \item Finally, we believe that applying segmentation to XAI methods can make the results to be consistent and easy to compare between XAI methods.
\end{itemize}

This paper's remainder is arranged to provide in the following order: related work, our proposed methods, experiments, and, ultimately, conclusions along with our future research directions.

\section{Related Work}
XAI methods can be categorized based on two factors. Firstly, the method can be intrinsic or post-hoc based on when the data is extracted. Secondly, the method can be either global or local based on the explanation's scope. Global models explain the complete, general model behavior and attempt to explain the whole logic of a model by inspecting the model's structures. Local models give explanations for a specific choice. For example, "Why the model have this prediction?". Global interpretability techniques lead users to trust a model. In reverse, local techniques lead users to believe in a prediction. Also, they try to identify each feature's contributions in the input towards a particular output \cite{DBLP:journals/corr/abs-1808-00033}. Post-hoc interpretation models can be applied to intrinsic models, but not fundamentally vice versa. LIME represents the Local Post-hoc approach \cite{Lime}, which is model-agnostic. In contrast, CAM represents the Local Intrinsic approach, which belongs to model-specific \cite{Cam}.
We also introduce the superpixel-based image segmentation method that we chose to use in this article 
\subsection{Segmentation Algorithms }
In the problem of image classification, the input image is of course a very important part. However, not every pixel in an image is meaningful. It would seem more intuitive to evaluate not only the perceptual but also the semantic meanings of an image created by locally grouping pixels. We get superpixels when we do this kind of local grouping of pixels on our pixel grid. It at the same time brings about computational efficiency benefits. It allows us to reduce the complexity of the image itself from hundreds of thousands of pixels down to just a few hundred superpixel. Each of these superpixels would then contain some sort of perceptual value and, ideally, semantics.
So, superpixels are becoming increasingly popular for use in computer vision applications. Superpixels provide a convenient original for calculating local image features\cite{slic}. The XAI algorithms in this field have segmented the image into superpixels and used the presence or absence of these superpixels as the interpretable representation \cite{anchors}. With this presentation, the image will be divided into regions. Each region will consist of several superpixels and will have certain meanings \cite{superpixels}. For example with the segmented image below, humans can easily see that the region consisting of superpixels number 11,12 and 20 will represent the hummingbird, where superpixels number 11 represent the bird’s beak, the bird’s head is the superpixels number 12,... 
    \begin{figure}[h]
        \captionsetup{justification=centering}
          \centering
          \includegraphics[height=0.5\linewidth,width=0.5\linewidth]{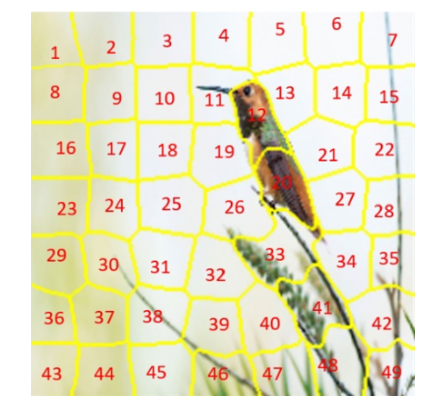}

        \caption{The segmented image for the input image (Figure 2a)}
        \label{fig:figure0}
    \end{figure}
    
The goal of the XAI methods in this case is to figure out which superpixels have the most influence on the model’s prediction. 
Some of the most commonly used superpixels methods are ETPS (Extended Topology Preserving Superpixels)\cite{etps}, SEEDS (Superpixels Extracted via Energy-Driven Sampling)\cite{seeds}, SLIC (Simple Linear Iterative Clustering)\cite{slic}, Quickshift \cite{quickshift},...
For superpixels to be useful they must be fast, easy to use, and produce high-quality segmentations. It is difficult to determine if segmentation is good or not because the definition of “good” often depends on the application. In this work, we experiment with the SLIC algorithm first and then with other algorithms. We will discuss the influence of segmentation algorithms later.
\subsection{Local Interpretable Model Agnostic Explanations}
     \textbf{Local Interpretable Model Agnostic Explanations (LIME)} is an XAI method that can explain any classifier or regressor's predictions assuredly by approximating it locally with an interpretable model \cite{Lime}. LIME intends to provide an easy to interpret method with local fidelity. The local fidelity means that the explanation for individual predictions should at least be locally faithful. In other words, it must correspond to how the model performs in the vicinity of the individual observation being predicted. The local fidelity does not imply global fidelity where the local context may not require globally essential features and vice versa. Due to this, even if a model has hundreds of variables globally, it could be the case that only a handful of variables directly relate to a local or individual prediction. LIME performs the steps below: 
     \begin{itemize}
         \item Generating new samples then gets their predictions using the original model.
         \item Weighing these new samples by the proximity to the instance being explained. 
     \end{itemize}
     Using the output probabilities from a given collection of samples that cover part of the input desired to be clarified, it then builds a linear model. Then, the surrogate model weights are used to measure the value of input features. Moreover, LIME is model-agnostic, so that it can be applied to any model of machine learning \cite{Lime}. 
     Figure\ref{fig:sfigure1b} is an example for LIME explanation with the Input image in Figure\ref{fig:sfigure1a} with model Resnet50.
\subsection{Class Activation Mapping}
     \textbf{Class Activation Mapping (CAM)} is a weighted activation map created for each input image \cite{Cam}. It utilizes a global average pooling (GAP) in CNNs. A class activation map for an appropriate category indicates the discriminative image regions used by CNN to identify that category. It is a locally intrinsic interpretable model that achieved by designing more justified model architectures \cite{DBLP:journals/corr/abs-1808-00033}. It explicitly allows CNNs to have exceptional localization ability despite being trained on image-level labels, enabling classification-trained CNNs to learn to produce object localization without using any bounding box annotations. CAM permits us to visualize the predicted class scores on any given image, highlighting the CNN's discriminative object parts. The CAM result shows a heatmap on the input image. This heatmap presents the impactful area of a given prediction \cite{Cam}.
\subsection{Gradient-weighted Class Activation Mapping}
    \textbf{Gradient-weighted Class Activation Mapping} (Grad-CAM) \cite{gradcam} is a generalized version of CAM. Grad-CAM uses the gradient information flowing into the last convolutional layer of the CNN to understand each neuron's decision of interest. Note that, to use CAM, the model must use a GAP layer followed by a fully connected softmax layer. This model architecture modification forces us to retrain the model. With a gradient approach, Grad-CAM can get the visualizations without changing the base model or retraining.
    
   The final feature convolutional map of the input image is activated for different class channels. In detail, every channel in the feature is weighed with the class gradient for that channel. The global average pooling over two dimensions \textit{(i,j)} for the gradient of respective class output for feature map is the spatial score of a specific class. Then, the resulting value is multiplied with the feature map along with the channel axis \textit{k}. While the resultant is pooled along its channel dimension. Hence, the spatial score map is of size \textit{i*j} which is normalized to positive region predictions using the nonlinear ReLU transformation. The class \textit{k}'s score correlates directly with the class-specific saliency map's importance, impacting the final prediction output.
     
\section{Proposed Methods}
\subsection{Motivation}
    In this section, we present the reason why we come up with the idea of SeCAM. In our previous work, we have applied LIME and CAM to explain the ResNet50 model. With the LIME method, we divided the image into 49 regions using the K-Means algorithm and calculated with the number of examples of 1000, which is the most appropriate number in this case. The results are shown in Figure \ref{fig:figure1}.
    
    \begin{figure}[h]
        \captionsetup{justification=centering}
            \begin{subfigure}{.33\linewidth}
              \centering
              \includegraphics[height=0.8\linewidth]{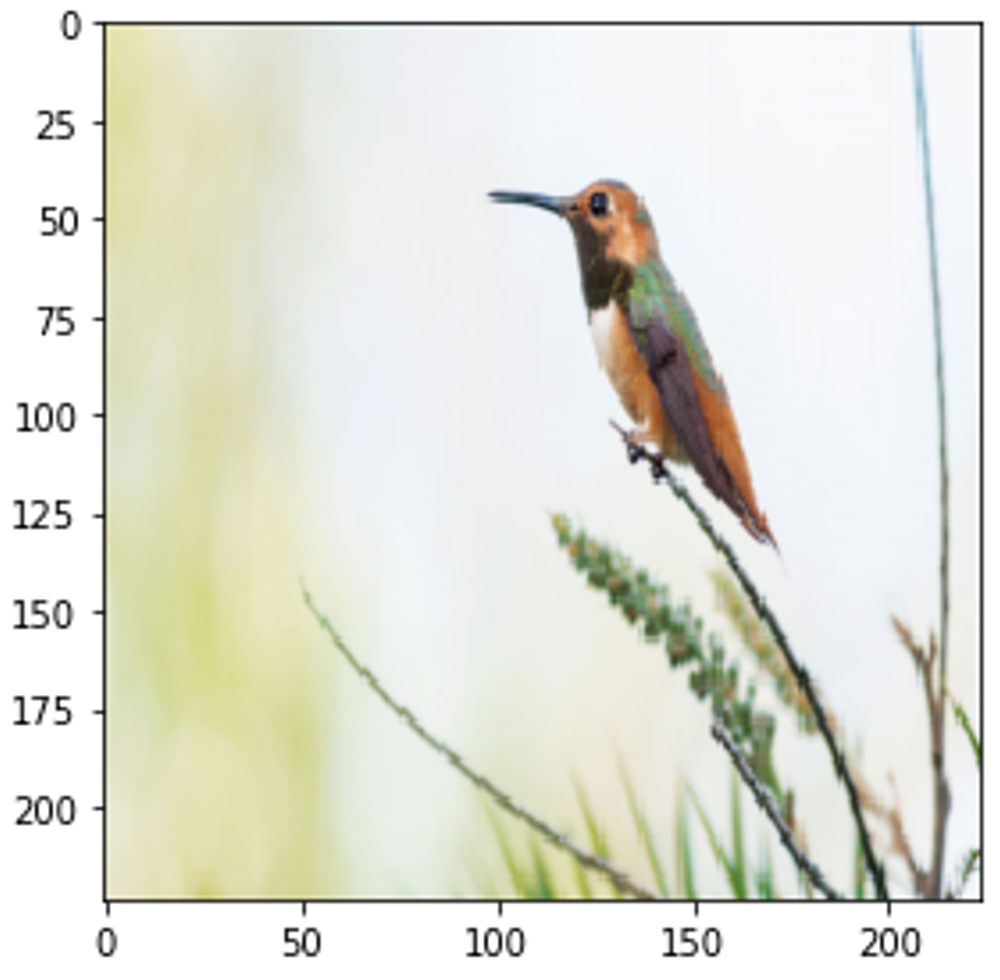}
              \caption{Input image}
              \label{fig:sfigure1a}
            \end{subfigure}%
            \begin{subfigure}{.33\linewidth}
              \centering
              \includegraphics[height=0.8\linewidth]{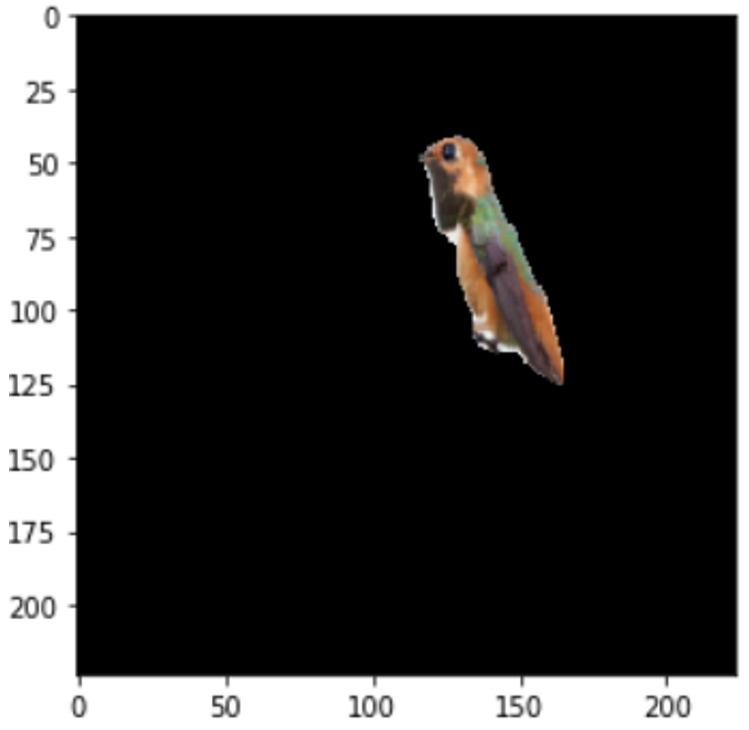}
              \caption{LIME}
              \label{fig:sfigure1b}
            \end{subfigure}%
            \begin{subfigure}{.33\linewidth}
              \centering
              \includegraphics[height=0.8\linewidth]{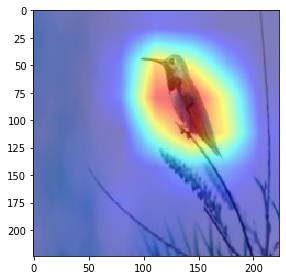}
              \caption{CAM}
              \label{fig:sfigure1c}
            \end{subfigure}
        \caption{The input image (a) and the results \\produced by LIME (b) and CAM (c). The label of the input image is Hummingbird.}
        \label{fig:figure1}
    \end{figure}
    
    The computation time of LIME and CAM are presented in Table \ref{table:1}.\\
    
    \begin{table}[h]
        \captionsetup{justification=centering,margin=1 em}
        \centering
        \caption{The computation time of LIME and CAM with the input image in Figure \ref{fig:sfigure1a}}
        \renewcommand{\arraystretch}{1.5}
            \begin{tabular}{|c|c|} 
            \hline
            \textbf{XAI methods} & \textbf{Running time (s)}  \\ 
            \hline
            \textbf{LIME}        & 235.761                \\ 
            \hline
            \textbf{CAM}          & 0.012                \\ 
            \hline
            \end{tabular}
        \label{table:1}
        \end{table}
    
    \begin{figure*}[hbt!]
        \captionsetup{justification=centering}
          \centering
          \includegraphics[width=\textwidth]{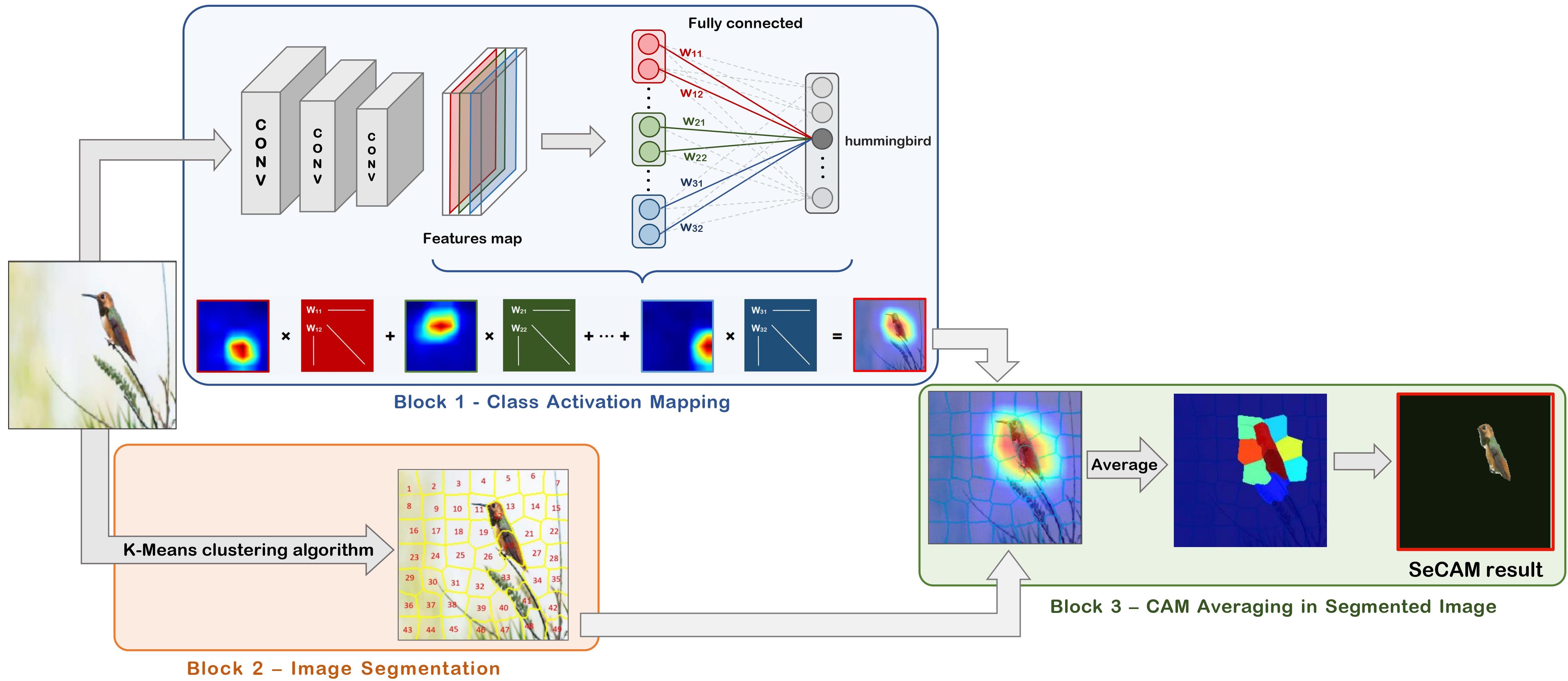}
        \caption{The architecture of Segmentation – Class Activation Mapping (SeCAM)}
        \label{fig:figure2}
    \end{figure*}
    
     The result in Table \ref{table:1} reveals that both LIME and CAM can yield the original image's regions that most affect the prediction. However, we find that LIME's explanation is resembling human explanations. The CAM heatmap area is too large, thus containing additional areas that do not have a decisive effect on the model's prediction, thereby reducing the explanation's reliability. As introduced in the Segmentation Algorithms section, with the results of LIME, we can see that the Resnet50 model rated the head and tail the most impact, more than the beak, but the InceptionV3 model shows that the head and the beak are the more important parts. Thus, by using superpixels, humans can see which regions will make more sense in making the model’s prediction instead of individual pixels like CAM. Nevertheless, LIME's computation time is too large, while CAM's computation time is completely superior with nearly 20000 times faster speed. The calculation time here is solely the time given for explaining. One of the prerequisites for using the CAM method is that at least one Global Average Pooling (GAP) layer exists in the model architecture \cite{Cam}. If the GAP layer is not already available in the model, then the model needs to be added with a GAP layer and retrains with all the data.
    
    To overcome the above problems, we propose a new method called \textit{Segmentation - Class Activation Mapping (SeCAM)} to improve LIME and CAM's disadvantages while preserving their advantages. Our proposed method produces a precise explanation of a predicted object like LIME but has a quick computation time of CAM. Furthermore, this method can be directly applied to any model with only an individual layer, followed by the last fully-connected softmax output layer. Thus, we do not have to add the GAP layer to the model or retrain the model anymore in the mentioned situation.

\subsection{Segmentation - Class Activation Mapping (SeCAM)}
    
    In this section, we describe our novel method - SeCAM in detail. As sketched in Figure 2, SeCAM consists of three blocks. In the first block, we initially apply the same procedure as the original CAM method, which identifies the image regions’ importance by projecting the output layer’s weights onto the convolutional feature maps [4]. For models without a GAP layer followed by a fully connected layer, we use the gradient information flowing into the last convolutional layer of the CNN (idea from Grad-CAM) and make some adjustments. So, our method SeCAM does not require models to have a GAP layer because it allows using any other layer such as a flatten layer to replace the GAP layer in the original CAM method, as shown in Figure 2. In block 2, we use a segmentation algorithm to segment the input image into superpixels. Results obtained from the previous two blocks are combined and compute the effect of each superpixel on the model’s prediction. In the following section, we will discuss more carefully about each block.
    
    \subsubsection{Block 1: Class Activation Mapping}
    
        For an input image, in the last convolutional layer, we get \(n\) feature maps. Let \(f_{k}(x, y)\) present the activation for unit \(k\) in feature maps at spatial location \((x, y)\). After a flattening class, each point represents a value of \(f_k(x, y)\). In the case of a pooling layer such as max or average pooling that turn multiple values \(f_k(x, y)\) (where \((x, y)\) belongs to a spatial location set A into a point, that point will represent multiple corresponding values for all spatial locations in the set A
        Therefore, for a class \(c\), we call the weights corresponding to the input to the softmax layer \(S_c = \sum_k\sum_{x, y}w_k^c(x, y)f_k(x, y)\) where \(w_k^c(x, y)\) is the weight corresponding to class \(c\) for unit \(k\) at \((x, y)\) location. In case the model already has a GAP layer followed by a fully connected softmax layer (Resnet50, InceptionV3,...), \(w_k^c(x,y)\)is taken directly from weight of \((x, y)\) in unit k corresponding to class \(c\). In the other cases,\(w_k^c(x,y)\) get the value of gradient via backpropagation:
        
        \begin{equation*}
            w_k^c(x,y) = \frac{\partial y^c}{\partial A^k} (x, y)
        \end{equation*}
        where \(A^k\) is the \(k^th\) feature map.
        In the other words, \(w_k^c(x, y)\) presents the importance of spatial element \((x, y)\) in unit \(k\) to class \(c\). After a softmax class, the output for class \(c\), \(P_c\) is determined by the Equation \ref{eq:exp}.
        
        \begin{equation}
            \frac{\exp{S_c}}{\sum_c\exp{S_c}}\label{eq:exp}
        \end{equation}. 
        
        Let \(M_c\) as the class activation map for class c, where each spatial element is given by Equation \ref{eq:1}.
        
        \begin{equation}
            M_c (x,y)=\sum_k w_k^c(x,y) f_k (x,y) \label{eq:1}
        \end{equation}
        
        Therefore,
        
        \begin{equation}
            \begin{aligned}
            S_c &=\sum_k \sum_{x,y} w_k^c (x,y) f_k (x,y) \\
            & = \sum_{x,y} \sum_k w_k^c(x,y)f_k (x,y) \\
            & =\sum_{x,y} M_c(x,y)\
           \end{aligned}
        \end{equation}\label{eq:2}
        
        Thus, \(M_c (x, y)\) indicates the importance of the activation at the spatial grid \((x, y)\), leading to the classification of input image to class \(c\).
        
    \subsubsection{Block 2: Image Segmentation}
        The Image Segmentation block runs parallel to the calculation of CAM values in Block 1. In this block, we split the input image into separate regions with similar coloring pixels. Hence, each region represents more meaningful and interpretable, and also carries more information than pixels. 
        
        Currently, there are many image segmentation algorithms. In the scope of this article, we use the K-Means algorithm to perform this division of images. More specifically, we use the simple linear iterative clustering (SLIC) algorithm, which is a particular case of K-means adapted to the task of generating superpixels. SLIC performs a local clustering of pixels in the 5-dimensional space defined by the L, a, b values of the CIELAB color space and the x, y pixel coordinates. We get high-quality segmentations with the SLIC algorithm with a low computational cost (SLIC achieves \(O(N)\) complexity) \cite{slic} [9]. With the SLIC algorithm, we can adjust the number of regions to be divided.
        The SLIC algorithm includes the following steps:
        
        \begin{itemize}
            \item Firstly, we initialize K cluster centers by sampling pixels at every grid interval \(S= \sqrt{N/K}\), where N is the number pixels of the input image.
            \item We move the centers to the new locations corresponding to the lowest gradient position 3 x 3 neighborhood. Image gradients are calculated as follow:
            \begin{equation*}
            \begin{split}
                G(x,y)= \|I(x+1,y)-I(x-1,y)\|^2 \\
                + \|I(x,y+1)-I(x,y-1)\|^2
            \end{split}
            \end{equation*}
            
            Where \(I(x,y)\) is the color vector in CIELAB color space corresponding to the pixel at position \((x,y)\), and $\|.\|$ is the \(L2\) norm.
            
            \item Similar to the K-means clustering \cite{}, we have a loop to update cluster centers and labels for all pixels. We repeat the following loop until convergence:
                \begin{itemize}
                    \item Update label for each pixel base on the nearest cluster center according to the distance measure Dsbetween a cluster center \(C_k = [l_k, a_k, b_k, x_k, y_k]^T\) and a pixel \( P_i=[l_i, a_i, b_i, x_i, y_i]^T\), is the sum of the \(lab\) distance and the \(xy\) plane distance normalized by the gri-d interval \(S\):
                    \[D_s = d_{lab} + \frac{m}{S}d_{xy}\]
                    
                    where:
                    \begin{itemize}
                        \item \(m\) is a parameter allowing us to control the density of a superpixel. The value of m can be in range \([1, 20]\). We choose 10 as the default value.
                        \item \(d_{lab}\) and \(d_{xy}\) are respectively the \(lab\) and the \(xy\) plane distances, defined as follow:
                        \begin{equation*}
                            d_{lab}=\sqrt{(l_k - l_i)^2 + (a_k - a_i)^2 + (b_k - b_i)^2}
                        \end{equation*}
                        \begin{equation*}
                            d_{xy}= \sqrt{(x_k - x_i)^2 + (y_k - y_i)^2}
                        \end{equation*}
                    \end{itemize}
                    
                    \item Compute a new center as the average labxyvector of all the pixels belonging to the cluster.

                \end{itemize}
                
                \item In the last step, if a few tray labels may remain, SLIC will enforce connectivity by relabeling disjoint segments with the labels of the largest neighboring cluster.

        \end{itemize}
    
    \subsubsection{Block 3: CAM Averaging in Segmented Image}
        Deriving the CAM values from Block 1 and the segmented image from Block 2, we average the values from the heatmap obtained in Block 1 for each region, called Segmentation Class Activation Mapping (SeCAM) value corresponding to that region. The SeCAM value for class \(c\) of region \(s\) is denoted \(M_c^s\) and is calculated by Equation \ref{eq:secamvalue}.
        
        \begin{equation}
            M_c^s = \frac{1}{|s|} \sum_{(x, y) \epsilon s}M_c (x, y)\label{eq:secamvalue}
        \end{equation}
        
        In which, \(|s|\) is the number of pixels in region \(s\). Thus, the \(M_c^s\) value represents each region's importance to the given prediction. The averaging ensures fairness between regions with different acreages and bypasses the requirement of adding a GAP class to the original model's architecture. When we take the average, each point in the region influences the \(M_c^s\) value. If the \(M_c^s\) value is maximized, the SeCAM ignores the effects of points with a smaller CAM value in one region.
        
       Finally, we select the regions that have the most significant impact on the prediction of the model, which are areas with the highest SeCAM values. There are two approaches to extract SeCAM’s explanation. The first way is to choose the number of regions with the most influence. The greater the number of regions, the broader the scope of explanation. The second way is to select the areas in which the value is above a given threshold of the SeCAM’s max value. The more extensive the threshold, the less the number of regions selected will be, and the smaller the explanation’s scope. We will discuss strategies for selecting the region later.

\section{Experimental Details}
    \begin{table}[ht]
        \captionsetup{justification=centering,margin=1.5 em}
        \centerline{\includegraphics[width=1\linewidth]{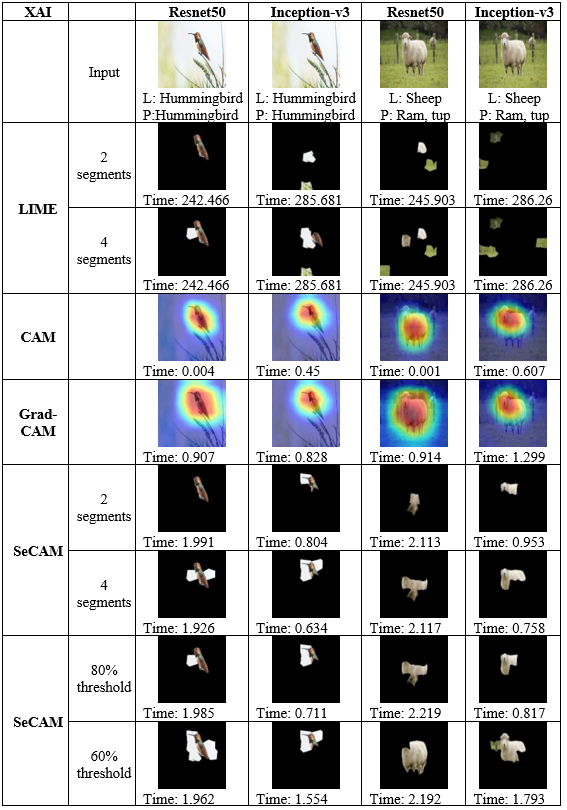}}
            \caption{The results and the running time of LIME, CAM, SeCAM method on the Hummingbird image and Sheep image ("L": Label, "P": Prediction).}
        \label{tab:tab2}
    \end{table}
\textbf{Experimental Setup}

   We conduct experiments for our SeCAM method, CAM, GradCAM, LIME on image from the dataset ILSVRC. We use models Resnet50, InceptionV3 and VGG16. In those models, Resnet50 and InceptionV3 have a GAP (Global Average Pooling) layer followed by a fully connected softmax layer, so we can use CAM with those two models easily. The VGG16 model is more complicated so CAM can’t be applied directly so we will use GradCAM instead.

\textbf{Quanlitative Results}   

We compare the explanation quality of LIME, CAM (GradCAM) and SeCAM on sample images from the ILSVRC dataset. The qualitative results for some images are shown in table II. The running time for each explanation is also included.

\textbf{Quantitative Results}

Since currently, to the extent of our knowledge, there are no accurate and recognized methods for comparative evaluation of XAI algorithms, we compare the precision between these results of SeCAM, LIME and CAM (GradCAM) on a human-grounded basis. Denoting G is the human-grounded bounding box and S is the bounding box of the explanation. We use the following evaluation metrics: 
\begin{itemize}
    \item \textbf{Intersection Over Union (IOU)} is the ratio between the overlapped area of two bounding boxes and their union area. IOU compares each bounding box produced by XAI methods to the ground truth.
     \begin{equation}
        IOU=\frac{S_{intersection}}{S_{union}}=\frac{S_{A\cap B}}{S_{A\cup B}} \label{eq:2}
    \end{equation}
    The IOU value varies from 0 to 1. The XAI method with the highest IOU value is the most accurate.
    
    \item \textbf{Energy-Based Pointing Game (EBPG)} evaluates the “accuracy” and variability of the XAI algorithms \cite{score-cam}. Extending the traditional pointing game, EBPG measures the fraction of its energy captured in the corresponding ground truth G.  \begin{equation}
            EBPG_S = \frac{\sum1_{(S\cap G)}\{x\}}{\sum1_{S}\{x\}} \label{eq:EBPG}
        \end{equation}
    In which, \(\sum1_{(S\cap G)}\{x\}\) is the number of points in both regions of S and G, and \(\sum1_{S}\{x\}\) is the number of points in S. So EBPS tells us what percentage of explanation's box S is in the human-grounded box G.
\end{itemize}

\section{Discussion}
\subsection{Computational resources}
    We used the 6 x RTX 2080 Ti GPU with Dual Xeon E5-2673 v3 CPU and 128Gb memory for experience. SeCAM is always the fastest and the slowest algorithm is LIME.
\subsection{SeCAM vs CAM}
    Through the Qualitative results in Table II. We find that the SeCAM results are not only closer, but also help us learn some insights of models. For example with the hummingbird image. The results of CAM (GradCAM) are heatmap regions related to the hummingbird bird and it is difficult to know which region has the most influence on the model's prediction when we  look at the explanation of model Resnet50. Meanwhile, the results of SeCAM can show the user the influence of each part of the hummingbird on the prediction results. Specifically, the hummingbird is divided into 3 main parts: the beak, the head and the body. The model resnet50 rated the body and head as the most important, while InceptionV3 took the beak and head. More specifically, the VGG16 model evaluates the beak as particularly important. However, with the GradCAM results, it is difficult to see that the head part also has a great influence on the prediction. This can be seen easier in the SeCAM results with 4 segments. 

\textbf{CAM obscure some important color parts}

During the experiment, we also found that, because the result of CAM or GradCAM is a heatmap, when overlaying the heatmap on the original image, it sometimes obscures some important parts. For example in Figure 3: In this case, the prediction of model InceptionV3 is Indigo bunting instead of coucal. We use CAM and SeCAM to explain why the model’s prediction is Indigo bunting. We found that the heatmap results of CAM obscured the characteristic blue color of the Indigo bunting species, while the SeCAM explanation showed more clearly the characteristic blue color area. With SeCAM we can see that the reason InceptionV3 predicts Indigo bunting is because of the similarity in color.

\textbf{CAM can not have a clear distinction between parts}

Also in the experimental process, in some cases, we find that the heatmap of CAM does not show humans which region affects the prediction the most, instead humans can only see that the heatmap contains the object and it is difficult to understand why the model makes that prediction. For example in Figure 4 and 5: In those cases, the InceptionV3 model predicts Input images 4.a and 5.d are Kite and Vulture instead of Coucal. If humans look at the explanation of CAM (Figure 4.b and 5.b), it's hard for humans to understand why the model predicted wrong. Meanwhile, based on SeCAM's explanation, we can see that the model evaluates the impact of the head is not high. So it is easier to understand why the model prediction Figure 4.a is a kite when we look at the SeCAM explanation.Especially with the 5.a image, the background with the surrounding dry branches also greatly affects the prediction of Vulture.

\subsection{SeCAM vs LIME}
When we compare SeCAM and LIME, the most obvious thing is that SeCAM and LIME's results are in similar form, indicating which superpixels have the most influence on the model's prediction results. However, the computation time of SeCAM is enormously faster with more than 100 seconds difference. Also, SeCAM ensures safety and stability when any image can give appropriate explanations to standard humans while LIME will be skewed in some instances, as shown in Table II, Figure 4 and Figure 5. So, we find that the quantitative results of SeCAM equal or surpass LIME.
\subsection{Effect of Segmentation Algorithms}
We also tried other fractional algorithms in addition to SLIC. Of course, with different algorithms, the way they segment the image is also different. However, XAI algorithms can still determine the exact affected area compared to human judgment. For example, in the Table III below, we have used two algorithms quickshift and SLIC, they produce different segmentation even though they have the same number of regions, but SeCAM can still select the correct areas of the image containing the object in both uses.
Therefore, the choice of different segmentation algorithms has no effect on model interpretation. We chose the SLIC algorithm because it is the easiest to understand, uses k-means clustering, and is the fastest implemented. At the same time, showing the segmentation areas of this algorithm is also friendly to the end-user, especially users who do not know much about technology.
\subsection{Comparition of XAI methods}
As introduced, comparing and evaluating XAI methods with each other is a big challenge. One of the reasons is that the results are out of sync. With LIME, SHAP results in perturbed images based on an segmentation algorithm, and CAM, GradCAM results in heatmaps. We believe that the idea of using segmentation integrated into heatmaps will not only improve the interpretation results, but also make comparisons between XAI methods easier. Because we can consider the explanation based on the influence of each part of the image instead of each pixel, comparing XAI methods and understanding the model behavior will be easier and save more time.

\section{Conclusions and Future Work}
    In this paper, we have introduced a novel method that explains the model’s prediction in the image classification problem, called Segmentation - Class Activation Mapping (SeCAM). Our method is developed based on these algorithms LIME, CAM, and GradCAM, the improved version of CAM. It combines the advantages of LIME’s explanation accuracy with the CAM’s calculation speed. Besides, SeCAM can represent the most significantly influential regions on the prediction and its impact value. These values can evaluate SeCAM’s accuracy based on humans grounded. We held a survey to see if the explanation of the new approach really got better. The results are extremely satisfactory. We experimented with many different image classification models on the ILSVRC dataset. In some cases, SeCAM gave the correct explanation and LIME gave rather absurd results. In explaining widely used image classification models, our method SeCAM results are more outstanding than other XAI methods such as LIME and CAM. There are a whole number of pathways of future work for us to explore with SeCAM. We recognize that our proposed method has limitations, and future academics and researchers should be aware of these and indeed interpret the material presented in this research within the context of the limitations. Firstly, there is a reasonably obvious limitation that the accuracy of SeCAM’s explanation depends too much on selecting the parameter for the selected segments or exact threshold level. Secondly, choosing the right algorithm for each model still has to be done manually. Currently, we are classifying into two main categories, which have multiple fully connected layers such as VGG16 and only one class fully connected layer, for example, ResNet50. The user will have to define the model's type in order to correctly apply the algorithm. We will try to find ways to automatically identify the model types in updated versions of the algorithm. Besides, we also find the inconvenience of the lack of a standard evaluation method for existing XAI methods; Therefore, we will also study to give a general evaluation method of the accuracy of different XAI algorithms in parallel with the development and improvement of the XAI algorithm.

\section{Acknowlegment}
We are grateful for the collaborative research environment provided by FPT Software Quy Nhon. We would like to express our special thanks of gratitude to Phong Nguyen for his sponsor and Prof. Takehisa Yairi from The University of Tokyo for his helpful support and discussions; Dr. Vinh Nguyen for his careful review. Finally, we would also like to acknowledge FSOFT AI Laboratory for providing us opportunities of incubating ideas in this project.

\mbox{~}
\clearpage
\newpage
\bibliographystyle{IEEEtran}
\bibliography{references} 

\begin{thebibliography}{10}
\providecommand{\url}[1]{#1}
\csname url@samestyle\endcsname
\providecommand{\newblock}{\relax}
\providecommand{\bibinfo}[2]{#2}
\providecommand{\BIBentrySTDinterwordspacing}{\spaceskip=0pt\relax}
\providecommand{\BIBentryALTinterwordstretchfactor}{4}
\providecommand{\BIBentryALTinterwordspacing}{\spaceskip=\fontdimen2\font plus
\BIBentryALTinterwordstretchfactor\fontdimen3\font minus
  \fontdimen4\font\relax}
\providecommand{\BIBforeignlanguage}[2]{{%
\expandafter\ifx\csname l@#1\endcsname\relax
\typeout{** WARNING: IEEEtran.bst: No hyphenation pattern has been}%
\typeout{** loaded for the language `#1'. Using the pattern for}%
\typeout{** the default language instead.}%
\else
\language=\csname l@#1\endcsname
\fi
#2}}
\providecommand{\BIBdecl}{\relax}
\BIBdecl

\bibitem{mohseni2020multidisciplinary}
\BIBentryALTinterwordspacing
S.~Mohseni, N.~Zarei, and E.~D. Ragan, ``A multidisciplinary survey and
  framework for design and evaluation of explainable ai systems,'' 2020.
  [Online]. Available: \url{https://arxiv.org/pdf/1811.11839.pdf}
\BIBentrySTDinterwordspacing

\bibitem{XAI}
A.~Adadi and M.~Berrada, ``Peeking inside the black-box: A survey on
  explainable artificial intelligence (xai),'' \emph{IEEE Access}, vol.~6, pp.
  52\,138--52\,160, 2018.

\bibitem{DBLP:journals/corr/abs-1808-00033}
\BIBentryALTinterwordspacing
M.~Du, N.~Liu, and X.~Hu, ``Techniques for interpretable machine learning,''
  \emph{CoRR}, vol. abs/1808.00033, 2018. [Online]. Available:
  \url{http://arxiv.org/abs/1808.00033}
\BIBentrySTDinterwordspacing

\bibitem{Lime}
\BIBentryALTinterwordspacing
M.~T. Ribeiro, S.~Singh, and C.~Guestrin, ``"why should {I} trust you?":
  Explaining the predictions of any classifier,'' \emph{CoRR}, vol.
  abs/1602.04938, 2016. [Online]. Available:
  \url{http://arxiv.org/abs/1602.04938}
\BIBentrySTDinterwordspacing

\bibitem{Shap}
\BIBentryALTinterwordspacing
S.~Lundberg and S.~Lee, ``A unified approach to interpreting model
  predictions,'' \emph{CoRR}, vol. abs/1705.07874, 2017. [Online]. Available:
  \url{http://arxiv.org/abs/1705.07874}
\BIBentrySTDinterwordspacing

\bibitem{Cam}
\BIBentryALTinterwordspacing
B.~Zhou, A.~Khosla, {\`{A}}.~Lapedriza, A.~Oliva, and A.~Torralba, ``Learning
  deep features for discriminative localization,'' \emph{CoRR}, vol.
  abs/1512.04150, 2015. [Online]. Available:
  \url{http://arxiv.org/abs/1512.04150}
\BIBentrySTDinterwordspacing

\bibitem{sise}
\BIBentryALTinterwordspacing
S.~Sattarzadeh, M.~Sudhakar, A.~Lem, S.~Mehryar, K.~N. Plataniotis, J.~Jang,
  H.~Kim, Y.~Jeong, S.~Lee, and K.~Bae, ``Explaining convolutional neural
  networks through attribution-based input sampling and block-wise feature
  aggregation,'' \emph{CoRR}, vol. abs/2010.00672, 2020. [Online]. Available:
  \url{https://arxiv.org/abs/2010.00672}
\BIBentrySTDinterwordspacing

\bibitem{saliencymap}
\BIBentryALTinterwordspacing
T.~N. Mundhenk, B.~Y. Chen, and G.~Friedland, ``Efficient saliency maps for
  explainable {AI},'' \emph{CoRR}, vol. abs/1911.11293, 2019. [Online].
  Available: \url{http://arxiv.org/abs/1911.11293}
\BIBentrySTDinterwordspacing

\bibitem{slic}
R.~Achanta, A.~Shaji, K.~Smith, A.~Lucchi, P.~Fua, and S.~Süsstrunk, ``Slic
  superpixels,'' \emph{Technical report, EPFL}, 06 2010.

\bibitem{anchors}
M.~T. Ribeiro, S.~Singh, and C.~Guestrin, ``Anchors: High-precision
  model-agnostic explanations,'' in \emph{AAAI Conference on Artificial
  Intelligence (AAAI)}, 2018.

\bibitem{superpixels}
\BIBentryALTinterwordspacing
D.~Stutz, A.~Hermans, and B.~Leibe, ``Superpixels: An evaluation of the
  state-of-the-art,'' \emph{CoRR}, vol. abs/1612.01601, 2016. [Online].
  Available: \url{http://arxiv.org/abs/1612.01601}
\BIBentrySTDinterwordspacing

\bibitem{etps}
\BIBentryALTinterwordspacing
S.~Zhang, C.~Li, S.~Qiu, C.~Gao, F.~Zhang, Z.~Du, and R.~Liu, ``Emmcnn: An
  etps-based multi-scale and multi-feature method using cnn for high spatial
  resolution image land-cover classification,'' \emph{Remote Sensing}, vol.~12,
  no.~1, 2020. [Online]. Available:
  \url{https://www.mdpi.com/2072-4292/12/1/66}
\BIBentrySTDinterwordspacing

\bibitem{seeds}
\BIBentryALTinterwordspacing
M.~V. den Bergh, X.~Boix, G.~Roig, and L.~V. Gool, ``{SEEDS:} superpixels
  extracted via energy-driven sampling,'' \emph{CoRR}, vol. abs/1309.3848,
  2013. [Online]. Available: \url{http://arxiv.org/abs/1309.3848}
\BIBentrySTDinterwordspacing

\bibitem{quickshift}
A.~Vedaldi and S.~Soatto, ``Quick shift and kernel methods for mode seeking,''
  in \emph{Computer Vision -- ECCV 2008}, D.~Forsyth, P.~Torr, and
  A.~Zisserman, Eds.\hskip 1em plus 0.5em minus 0.4em\relax Berlin, Heidelberg:
  Springer Berlin Heidelberg, 2008, pp. 705--718.

\bibitem{gradcam}
\BIBentryALTinterwordspacing
R.~R. Selvaraju, M.~Cogswell, A.~Das, R.~Vedantam, D.~Parikh, and D.~Batra,
  ``Grad-cam: Visual explanations from deep networks via gradient-based
  localization,'' \emph{International Journal of Computer Vision}, vol. 128,
  no.~2, p. 336–359, Oct 2019. [Online]. Available:
  \url{http://dx.doi.org/10.1007/s11263-019-01228-7}
\BIBentrySTDinterwordspacing

\bibitem{score-cam}
\BIBentryALTinterwordspacing
H.~Wang, M.~Du, F.~Yang, and Z.~Zhang, ``Score-cam: Improved visual
  explanations via score-weighted class activation mapping,'' \emph{CoRR}, vol.
  abs/1910.01279, 2019. [Online]. Available:
  \url{http://arxiv.org/abs/1910.01279}
\BIBentrySTDinterwordspacing

\end{thebibliography}

\vfill\break

\section*{Appendix}
\subsection{Result images in experiment}
In the experiment, we have applied SeCAM, CAM and LIME on various images from the ILSVRC dataset. In the Table \ref{tab:tab3}, we present three examples of results with different XAI methods' parameters.

  \begin{table}[ht]
  \captionsetup{justification=centering}
        \centerline{\includegraphics[width=1\linewidth]{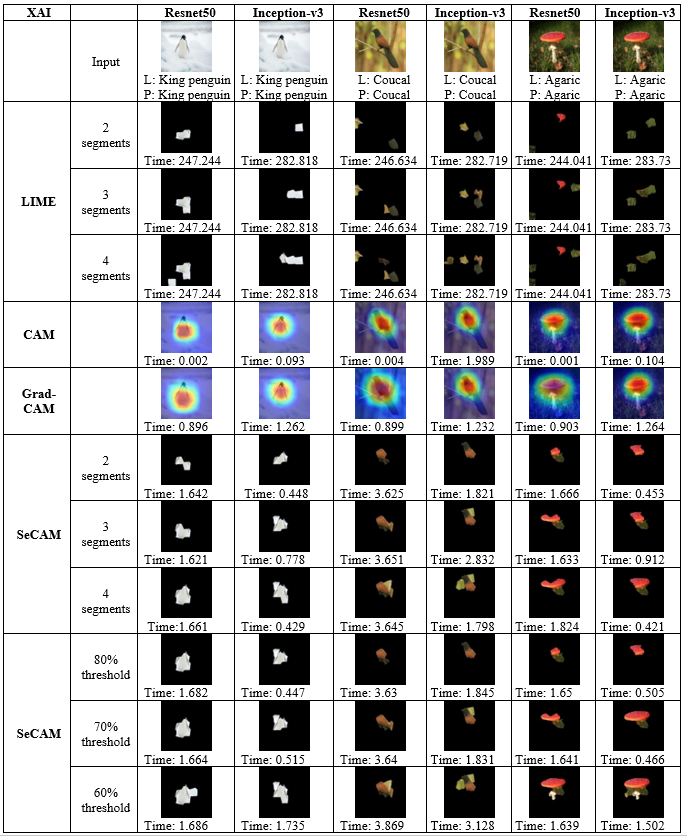}}
            \caption{The results of LIME, CAM, SeCAM on images of the ILSVRC dataset.}
        \label{tab:tab3}
    \end{table}
\end{document}